\documentclass[letterpaper]{article} 
\usepackage{aaai24}  
\usepackage{times}  
\usepackage{helvet}  
\usepackage{courier}  
\usepackage[hyphens]{url}  
\usepackage{graphicx} 
\def\UrlFont{\rm}  
\usepackage{natbib}  
\usepackage{caption} 
\usepackage{algorithm}
\usepackage{algorithmic}

\usepackage{multirow}
\usepackage{amsmath}
\usepackage{xcolor}
\usepackage{makecell}

\usepackage{newfloat}
\usepackage{listings}
\DeclareCaptionStyle{ruled}{labelfont=normalfont,labelsep=colon,strut=off} 
\floatstyle{ruled}
\newfloat{listing}{tb}{lst}{}
\floatname{listing}{Listing}
\title{Efficient Toxic Content Detection by Bootstrapping and Distilling \\ Large Language Models}
\author{
    Jiang Zhang\textsuperscript{\rm 1}\thanks{This work was done when the author was an intern at Amazon.},
    Qiong Wu\textsuperscript{\rm 2},
    Yiming Xu\textsuperscript{\rm 2},
    Cheng Cao\textsuperscript{\rm 2},
    Zheng Du\textsuperscript{\rm 2},
    Konstantinos Psounis\textsuperscript{\rm 1}
}
\affiliations{
    \textsuperscript{\rm 1}University of Southern California,
    Los Angeles, CA, USA,  \qquad \textsuperscript{\rm 2}Amazon.com, Inc., Bellevue, WA, USA\\
    \{jiangzha,kpsounis\}@usc.edu, \qquad \{qionggwu,ymxu,chengcao,zhengdu\}@amazon.com
    
}

\begin{document}
\maketitle

\begin{abstract}
Toxic content detection is crucial for online services to remove inappropriate content that violates community standards. To automate the detection process, prior works have proposed varieties of machine learning (ML) approaches to train Language Models (LMs) for toxic content detection. However, both their accuracy and transferability across datasets are limited. Recently, Large Language Models (LLMs) have shown promise in toxic content detection due to their superior zero-shot and few-shot in-context learning ability as well as broad transferability on ML tasks.
However, efficiently designing prompts for LLMs remains challenging. Moreover, the high run-time cost of LLMs may hinder their deployments in production. To address these challenges, in this work, we propose \textbf{BD-LLM}, a novel and efficient approach to \textbf{B}ootstrapping and \textbf{D}istilling \textbf{LLM}s for toxic content detection. 
Specifically, we design a novel prompting method named Decision-Tree-of-Thought (DToT) to bootstrap LLMs' detection performance and extract high-quality rationales. DToT can automatically select more fine-grained context to re-prompt LLMs when their responses lack confidence. Additionally, we use the rationales extracted via DToT to fine-tune student LMs. Our experimental results on various datasets demonstrate that DToT can improve the accuracy of LLMs by up to 4.6\%. Furthermore, student LMs fine-tuned with rationales extracted via DToT outperform baselines on all datasets with up to 16.9\% accuracy improvement, while being more than 60$\times$ smaller than conventional LLMs. Finally, we observe that student LMs fine-tuned with rationales exhibit better cross-dataset transferability.

\end{abstract}

\section{Introduction}
Toxic content detection is important for online services to protect users from harmful and offensive content, ensuring a safer and more positive user experience. Common toxic content categories include hate speech, biased content, sexual content, violent content, bullying content, etc. Due to the massive amount of content on the Internet, it is impractical to manually check the toxicity of each content. Hence, machine learning (ML) solutions based on supervised learning have been widely applied to automate the toxic content detection process, where Language Models (LMs) fine-tuned on a task-specific dataset achieve the state-of-the-art (SOTA) performance \cite{caselli2020hatebert,kim2022generalizable}.

However, existing supervised learning ML solutions face three challenges. First, they require training data with labels, which are non-trivial to obtain for toxic content detection tasks due to the lack of standard definitions, especially for implicit toxic content. Second, the fine-tuned LMs may overfit the training dataset, which limits the transferability to other datasets. Lastly, they typically can only predict binary labels without detailed reasoning. 

To handle the above challenges, the recently emerging Large Language Models (LLMs) have been leveraged to detect toxic content \cite{wang2022toxicity,zhang2023interpretable}, due to their superior zero-shot and few-shot in-context learning performance and transferability. Existing works on LLMs for toxic content detection focus on designing novel prompting approaches to enhance the performance of LLMs. However, their performance relies heavily on the quality of prompts, which are non-trivial to design. 
Moreover, deploying LLMs for toxic content detection in production can incur both high run-time cost and high latency, especially when the number of tokens in the prompt is large (e.g. for in-context few-shot learning). 


\begin{figure*}[t]
\centering
\includegraphics[width=0.80\textwidth]{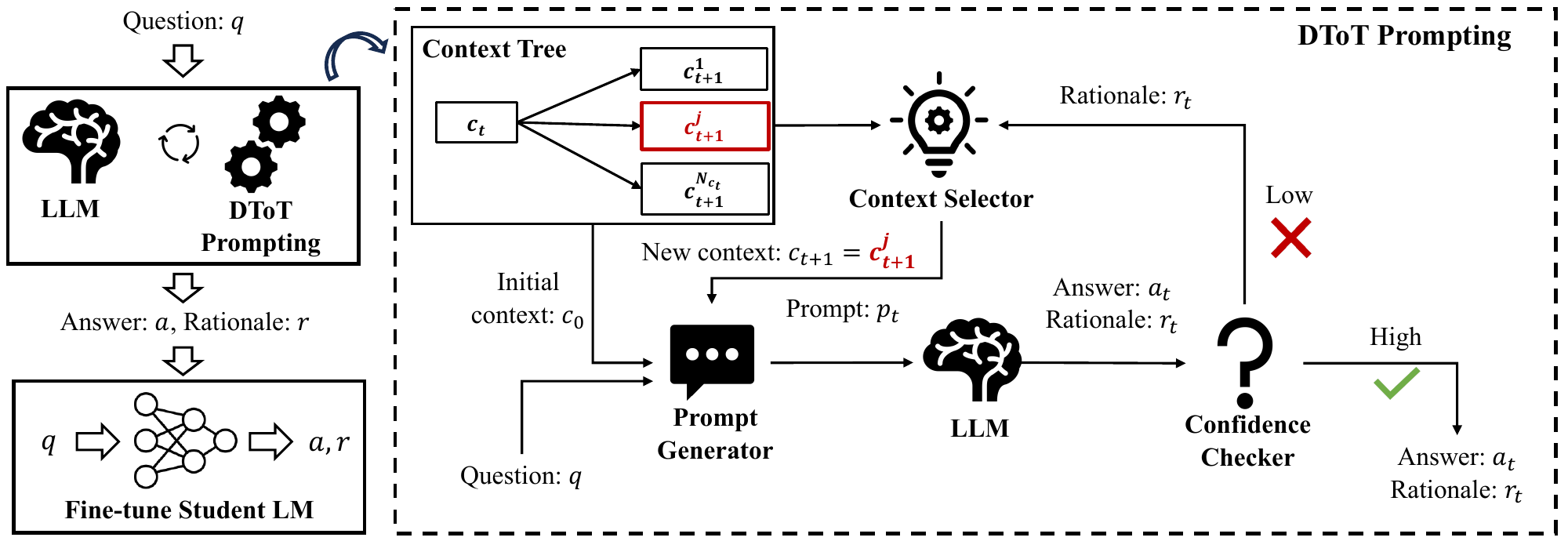}
\vspace{-0.08in}
\caption{Overall workflow of the proposed BD-LLM. Given question $q$, it first bootstraps the LLM via DToT prompting to extract answer $a$ and rationale $r$ with high-confidence. Then, it uses $q$ as input and $(a,r)$ as output to fine-tune the student LM.}
\label{fig:overall}
\vspace{-0.17in}
\end{figure*}

Motivated by the above, in this paper, we propose \textbf{BD-LLM}, a novel and efficient approach to \textbf{B}ootstrapping and \textbf{D}istilling \textbf{LLM}s for toxic content detection (as shown in Figure \ref{fig:overall}). We first design a novel prompting approach called Decision-Tree-of-Thought (DToT) to improve the zero-shot and few-shot in-context learning  performance of LLMs as well as extract better rationales. At a high level, DToT works by iteratively selecting more fine-grained context to re-prompt LLMs whenever they have low-confident answers. It automatically decides when to select more fine-grained context for re-prompting and which type of fine-grained context should be selected. Second, we propose to fine-tune a student LM with smaller model size to predict both labels and rationales extracted via DToT.

We evaluate the proposed approach on three public datasets and one Amazon internal dataset. Experimental results demonstrate that employing DToT prompting consistently leads to improved zero-shot and few-shot in-context learning performance on all datasets, by up to 4.6\% higher accuracy. Furthermore, we demonstrate that fine-tuning student LMs with DToT-extracted rationales results in up to 16.9\% accuracy improvement compared with baselines.
Meanwhile, these student LMs have model size more than 60$\times$ smaller than conventional LLMs. Finally, we observe that the cross-dataset transferability of student LMs fine-tuned with rationale is significantly improved.

Our contributions are summarised as follows:
\begin{itemize}
    \item We propose BD-LLM, an efficient approach for toxic content detection, which leverages LLMs strength but also reduces their complexity via bootstrapping and distillation. This is the first-of-its-kind study on toxic content detection.
    \item To bootstrap LLMs, we design a novel prompting method named DToT, which selectively re-prompts LLMs with more fine-grained context to boost their detection performance and extract high-quality rationales.
    \item To distill LLMs into smaller student LMs, we fine-tune the student LMs to predict both labels and rationales extracted via DToT.
    \item We evaluate the proposed solution on four datasets and demonstrate that DToT can improve the toxic content detection accuracy of LLMs by up to 4.6\% and student LMs fine-tuned with DToT-extracted rationales achieve the SOTA performance with up to 16.9\% accuracy improvements and more than 60$\times$ smaller size.
\end{itemize}


\vspace{-0.08in}
\section{Related Work}
\subsection{Toxic Content Detection}
\vspace{-0.03in}
Prior works on toxic content detection can be categorized into two types.
One type of research works focuses on creating benchmark datasets for toxic content detection, either by crowdsourcing and annotating human-written text \cite{ye2023noisyhate, sap2019social,vidgen2020learning}, or leveraging ML-based approaches to generate high-quality toxic dataset in a scalable way \cite{hartvigsen2022toxigen}. 
Another type of works proposes novel approaches to fine-tune LMs on toxic dataset. \citeauthor{caselli2020hatebert} \shortcite{caselli2020hatebert} propose HateBERT, a pre-trained BERT model for abusive language detection, which significantly outperforms the original BERT model. \citeauthor{kim2022generalizable} \shortcite{kim2022generalizable} propose a novel contrastive learning method to improve the cross-dataset performance of HateBert. Most recently, researchers have started to use LLMs to detect toxic content. \citeauthor{wang2022toxicity} \shortcite{wang2022toxicity} design a generative prompt-based inference method to leverage LLMs for toxic content detection. \citeauthor{zhang2023interpretable} \shortcite{zhang2023interpretable} propose an interpretable, unified, language checking method (UniLC) to enhance the few-shot in-context learning capabilities of LLMs for misinformation, stereotypes, and hate
speech detection. Compared with these works, our work not only proposes a novel and orthogonal prompting approach that improves the zero-shot/few-shot performance, but also extracts and  distills rationales into a smaller but more effective student LM for toxic context detection.

\vspace{-0.06in}
\subsection{Prompting LLMs}
\vspace{-0.03in}
LLMs such as GhatGPT \cite{chatgpt} and Llama \cite{touvron2023llama} have demonstrated superior zero-shot/few-shot in-context learning capabilities and generalizability on a variety of language tasks without fine-tuning \cite{kojima2022large}. However, their performance is heavily related to the quality of input prompts \cite{zhao2023survey}. Hence, varieties of prompting approaches have been proposed to improve the quality and robustness of LLMs' response \cite{wei2022chain,yao2023tree,wang2022self,luo2023sail,wang2023large,chen2023zara}. \citeauthor{wei2022chain}\shortcite{wei2022chain} and \citeauthor{wang2022self}\shortcite{wang2022self} propose Chain-of-Thought (CoT) and self-consistent CoT, which prompts LLMs step by step to improve LLMs' performance on reasoning tasks. \citeauthor{yao2023tree}\shortcite{yao2023tree} generalizes CoT into Tree-of-Thought (ToT), which enables LLMs to explore multiple intermediate steps for complex problem solving tasks. Different from CoT and ToT which are suitable for step-by-step reasoning problems, the proposed DToT in this work is designed for classification problems with a novel confidence checker and context selector, which iteratively searches and injects more fine-grained context into prompts to enhance the classification confidence of LLMs. Other works have proposed to leverage the in-context learning capability of LLMs, by augmenting the prompts with demonstrations \cite{wang2023large}, rationales \cite{chen2023zara} or grounded information \cite{luo2023sail,zhang2023interpretable}. Note that our DToT prompting is orthogonal to existing in-context learning methods for LLMs (see Section 3.2).

\vspace{-0.06in}
\subsection{Distilling LLMs}
\vspace{-0.03in}
Some recent works also tackle the problem by distilling LLMs into smaller LMs for domain-specific tasks \cite{magister2022teaching,hsieh2023distilling,wang2022pinto,wang2023scott}. 
\citeauthor{wang2022pinto} \shortcite{wang2022pinto} propose PINTO, which uses LLMs' rationales as context of a small LM to improve its performance on reasoning tasks. \citeauthor{wang2023scott} \shortcite{wang2023scott} further propose SCOTT, a faithful knowledge distillation method that elicits self-consistent rationales from LLMs to fine-tune a small while more faithful LM for open-domain question-answering tasks. Both \citeauthor{magister2022teaching} \shortcite{magister2022teaching} and \citeauthor{hsieh2023distilling} \shortcite{hsieh2023distilling} demonstrate that LLMs can be distilled into smaller but more effective LMs by fine-tuning with both answers and rationales on commonsense reasoning and arithmetic tasks. Different from these works, our work focus on the domain of toxic content detection. Moreover, we propose DToT that extracts better rationales from LLMs and demonstrate that using DToT-extracted rationales further improves the effectiveness of distillation for toxic content detection.
\section{Approach}
Figure \ref{fig:overall} demonstrates the overall workflow of BD-LLM, which consists of two separate steps. Specifically, in the first step, we design DToT prompting to iteratively extract high-confidence answers $a$ and rationales $r$ from LLM given input question $q$.~\footnote{Note that for toxic content detection, an answer is either `Yes' or `No', while a rationale consists of one or more sentences.} In the second step, we conduct rationale distillation by fine-tuning a student LM to generate both $a$ and $r$ given input $q$. We describe the details of each step below.

\vspace{-0.05in}
\subsection{DToT Prompting}
At a high level, DToT prompting iteratively selects more fine-grained context and re-prompts the LLM when they output responses with low confidence. Two challenges in designing DToT prompting are: \textit{1) how to decide whether the response of LLM is confident or not}; \textit{2) how to decide which type of fine-grained context should be selected to re-prompt the LLM\footnote{Since there are many different types of toxic content and for each of them we need to use the right context (i.e. part of the prompt that specifies the criteria to call certain content as toxic), there is a need to automatically select the appropriate context for prompting.}}. To address the first challenge, we design a confidence checker to measure the self-confidence score of the LLM's answer. To handle the second challenge, we design a context selector to select appropriate fine-grained context from a context tree based on the LLM's rationale. 

In total, DToT prompting consists of four modules: 1) confidence checker; 2) context tree; 3) context selector; 4) prompt generator, which can be used for both \textit{black-box} and \textit{white-box} LLMs. Note that for \textit{black-box} LLMs, we assume that we only have access to their output responses (e.g. ChatGPT). By contrast, for \textit{white-box} LLMs, we assume that we can also have access to their model parameters (e.g. open-source LLMs). The detailed design of these modules and the end-to-end workflow are presented below.

\vspace{-0.05in}
\subsubsection{Confidence Checker.} This module is designed to measure the self-confidence score of LLM's response, which is defined as $s_t^{conf}$ for iteration step $t$ . Specifically, for the \textit{black-box} LLM, the confidence checker uses the toxicity rating of LLM (defined as $s_t^{toxi}\in[0,100]$) to calculate the confidence score. If $s_t^{toxi}$ is above a maximal threshold $s_{h}$ or below a minimal threshold $s_{l}$, we consider the answer as confident, vice versa. Note that to obtain $s_t^{toxi}$, we explicitly require the LLM to output the toxicity rating in the prompt $p_t$. For the \textit{white-box} LLM (e.g. any open-source LLMs), the confidence checker will leverage the output logits of LLM to calculate the probability of generating answers $a_t$ conditional on the prompt $p_t$. This probability will be used as the confidence measurement (i.e. $s_t^{conf}=\mathrm{Pr}[a_t|p_t]$). Formally, we define $s_t^{conf}$ as: 
\begin{equation} \label{eq:score} 
s_t^{conf}= \begin{cases}
    \mathbf{1}[s_t^{toxi}\notin(s_{l},s_{h})], & \text{for \textit{black-box} LLMs}\\ 
    \mathrm{Pr}[a_t|p_t],   & \text{for \textit{white-box} LLMs}
\end{cases}
\end{equation}
where $\mathbf{1}[\cdot]$ is an indicator function, and $s_{l}$ and $s_{h}$ are two adjustable thresholds (see Appendix \ref{appendix:param} for selecting $s_{l}$, $s_{h}$).

Suppose the confidence checker has measured the self-confidence score $s_t^{conf}$ of LLM's answer $a_t$. Then, if $s_t^{conf}$ is higher than a threshold $s_{\delta}$, the checker will consider $a_t$ as a confident answer. Otherwise, $a_t$ is considered to be unconfident. Formally, the output of the confidence checker $D_{check}$ is defined as:
\begin{equation} \label{eq:conf} 
D_{check}(s_t^{conf}) = \begin{cases}
    \mathrm{Unconfident,} & \text{if}~s_t^{conf}\in[0,s_{\delta})\\ 
    \mathrm{Confident,} & \text{if}~s_t^{conf}\in[s_{\delta},1]
\end{cases}
\end{equation}
where $s_{\delta}$ is an adjustable threshold (see Appendix \ref{appendix:param} for selecting $s_{\delta}$).

\vspace{-0.05in}
\subsubsection{Context Tree.} 
\begin{figure}
    \centering
    \includegraphics[width=0.85\linewidth]{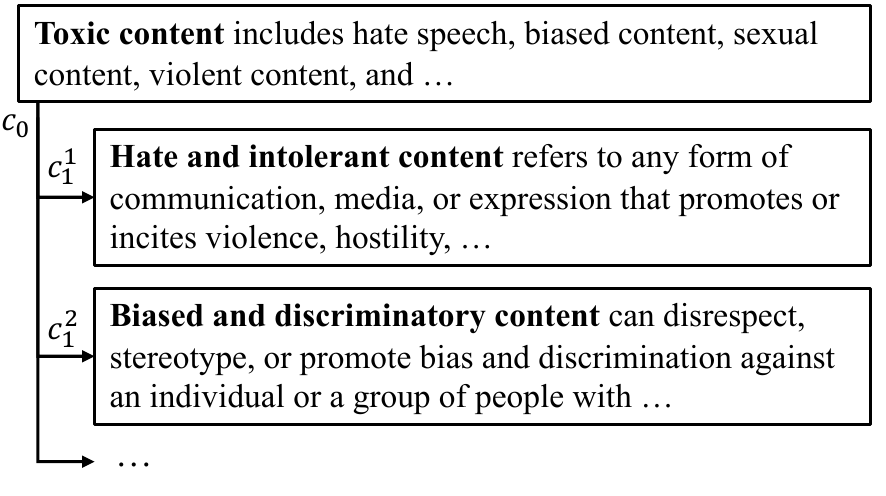}
    \vspace{-0.16in}
    \caption{An example of context tree.}
    \vspace{-0.15in}
    \label{fig:tree}
\end{figure}

Before introducing the context selector module, we first provide the definition of context tree. Suppose the universe of context is represented as set $\mathcal{C}$. We define the context tree as $T_{c}:\mathcal{C}\rightarrow \textsc{List}[\mathcal{C}]$, which is a mapping from a parent-node context $c_t\in\mathcal{C}$ to the list of its child-node contexts $[c_{t+1}^{1},...,c_{t+1}^{N_{c_t}}]$, where $N_{c_t}$ is the total number of child nodes of $c_t$, and $c_{t+1}^{j}\in\mathcal{C}$ is the $j$-th child-node context of $c_t$. 
Moreover, each child-node context's category is designed to be a subcategory of its parent-node context's category, such that the child-node context is more fine-grained than parent-node context. For instance, suppose that $c_{0}$ provides the definition of toxic content, which includes hate speech, sexual content, etc. Then, $c_{1}^{1}$ can be a more fine-grained definition for hate speech, and $c_{1}^{2}$ can be a more fine-grained definition for biased content (see Figure \ref{fig:tree}).

\vspace{-0.05in}
\subsubsection{Context Selector.} Now we introduce the context selector module, which takes as input the rationale $r_t$ and the context $c_t$ in prompt at step $t$, and a context tree $T_{c}$, to select a more fine-grained context $c_{t+1}$ for step $t+1$.  Whenever an answer $a_t$ at iteration step $t$ is considered as unconfident by the confidence checker, the context selector will select new context from the context tree for re-prompting LLMs.
Specifically, it measures the relevance between rationale $r_t$ and each child-node context $c_{t+1}^{j}$, and then selects the most relevant one $c_{t+1}=c_{t+1}^{j^*}$. 
Formally, we define the output of context selector as:
\begin{equation} \label{eq:select} 
D_{select}(r_t)\!=\!j^{*}\!=\arg\!\!\max_{1\leq j\leq N_{c_t}}\!{s_{t}^{rele,j}(r_t,c_{t+1}^{j})},
\vspace{-0.03in}
\end{equation}
where $[c_{t+1}^{1},...,c_{t+1}^{N_{c_t}}]=T_c(c_t)$, and $s_{t}^{rele,j}$ is the relevance score between $r_t$ and $c_{t+1}^j$.

Furthermore, to calculate $s_{t}^{rele,j}$ for LLMs which can generate relatively high-quality rationales (mostly for \textit{black-box} LLMs like ChatGPT), we use a classification prompt to ask LLM which of these child-node context categories are most relevant to the rationale $r_t$. We denote the answer from the LLM as $\mathrm{Class}(r_t)$. 
By contrast, for \textit{white-box} LLMs which cannot generate rationales with decent quality, we construct a set of candidate rationales as $r_t = [r_{t}^{1},...,r_{t}^{N_{c_t}}]$, where each $r_{t}[j]=r_{t}^{j}$ is relevant to only one of the child-node contexts $c_{t+1}^{j}$. Since we have access to the output logits of the LLM, we measure the probability of generating each candidate rationale conditional on the prompt $p_t$ (i.e., $\mathrm{Pr}[r_{t}[j]|p_t]$) as the relevance score. Formally, we define $s_{t}^{rele,j}$ as:
\begin{equation} \label{eq:select_score} 
s_{t}^{rele,j} = \begin{cases}
    \mathbf{1}[j=Class(r_t)], & \text{for \textit{black-box} LLMs}\\ 
    \mathrm{Pr}[r_{t}[j]|p_t],   & \text{for \textit{white-box} LLMs}
\end{cases}
\vspace{-0.03in}
\end{equation}
where $\mathbf{1}[\cdot]$ is an indicator function.

\vspace{-.05in}
\subsubsection{Prompt Generator.} The Prompt Generator module $P$ generates an input prompt $p_t$ for LLM $M$ based on the question $q$, and the context $c_t$ at iteration step $t$. Note that it will modify the question $q$ with the change of context. For example, the initial question based on $c_0$ is designed to ask whether a statement contains toxic content. Suppose $c_1$ is selected to provide context related to hate speech. Then, the question will be modified to ask whether the statement contains hate speech, which is a specific category of toxic content. We provide the prompt templates in Appendix \ref{appendix:prompt}.

\vspace{-.05in}
\subsubsection{End-to-end Workflow.}
Algorithm \ref{alg:algorithm} illustrates the end-to-end workflow of DToT prompting, where the input contains LLM $M$, question $q$, initial context $c_0$. At each iteration step $t$ (line 2), it first generates prompt $p_t$ via prompt generator $P$ and gets LLM $M$'s output (lines 4-10). Next, it calculates the confidence score of LLMs's answer $s_t^{conf}$ and decides whether it is confident via the confidence checker $D_{check}$ (lines 12-13). For unconfident answers, it will get a list of new candidate contexts from context tree $T_c$ (line 15), calculate the relevance score between LLM's rationale and each one of the new context in the list (lines 17-24), and select the most relevant new context (lines 25-26). Lastly, it terminates when the maximal iteration step $T$ is reached (line 2) or the LLM's answer is confident (line 29).

\begin{algorithm}[!h]
\caption{DToT Prompting}
\label{alg:algorithm}
\textbf{Input}: Question $q$, initial context $c_0$, LLM $M$, thresholds $s_{l}$, $s_{h}$, and $s_{\delta}$, maximal iteration step $T$.\\
\textbf{Output}: Answer $a$, rationale $r$.
\begin{algorithmic}[1] 
\STATE Define current step as $t=0$.
\WHILE{$t < T$}
\STATE \textcolor{blue}{// Generate prompt and get response}
\STATE Generate input prompt: $p_t=P(q, c_t, M)$.
\IF {$M$ is black-box}
\STATE Get LLM output: $(a_t, s_t^{toxi}, r_t) = M(p_t)$.
\ELSE
\STATE Get LLM output: $(a_t, r_t) = M(p_t)$.
\ENDIF
\STATE \textcolor{blue}{// Check the confidence of answer}
\STATE Calculate confidence score $s_t^{conf}$ by Eq. (\ref{eq:score}).
\IF {$D_{check}(s_t^{conf})=\mathrm{Confidence}$:}
\STATE \textcolor{blue}{// Get candidate new contexts from context tree}
\STATE Get the new context list: $[c_{t+1}^{j}]_{j=1}^{N_{c_t}}=T_c(c_t)$.
\STATE \textcolor{blue}{// Select new context}
\IF {$M$ is black-box}
\STATE Let $M$ get the rationale class: $\mathrm{Class}(r_t)$.  
\ELSE
\STATE Let $r_t$ be the candidate rationale list: $[r_{t,j}]_{j=1}^{N_{c_t}}$.
\STATE Calculate relevance score $[s_{t}^{rele,j}]_{j=1}^{N_{c_t}}$ by Eq. (\ref{eq:select_score}).
\ENDIF
\STATE Calculate $j^{*}=D_{select}(r_t)$ by Eq. (\ref{eq:select}).
\STATE Set new context $c_{t+1}$ as $c_{t+1}^{j^{*}}$ and  $t = t + 1$.
\ELSE
\STATE Exit.
\ENDIF
\ENDWHILE
\STATE \textbf{return} $a=a_t, r=r_t$
\end{algorithmic}
\end{algorithm}

\vspace{-0.05in}
\subsection{Augmented DToT Prompting}
\label{subsec:dtot_aug}
So far, we present how DToT prompting works under the zero-shot learning setup. Considering that DToT prompting is orthogonal to few-shot learning, we propose two augmented versions of DToT prompting below.

\vspace{-0.05in}
\subsubsection{DToT+FS.} The approach combines DToT prompting with the vanilla few-shot in-context learning, which adds a few demonstrations in the prompt. Recent work \cite{wang2023large} has shown that a few demonstrations can effectively improve the in-context learning performance of LLMs. Moreover, motivated by prior work \cite{liu2021makes}, we select $K$ positive statements and $K$ negative statements that are most semantically similar to the input statement as demonstrations from a development set. 

\vspace{-0.05in}
\subsubsection{DToT+FS+R.} Recent works \cite{sun2022recitation,zhang2023interpretable} have demonstrated that rationales or facts in prompts serve as grounded information to further enhance the few-shot in-context learning capability of LLMs. Therefore, on top of DToT+FS approach, we include the rationale for each demonstration in the prompt.

\subsection{Rationale Distillation}
After obtaining the answers and rationales from LLMs via DToT, we can distill LLMs into smaller student LMs, which can predict both answers and rationales. We describe the details of rationale distillation under two scenarios below.

\vspace{-0.02in}
\subsubsection{Distillation without Labels.} Assume that we do not have ground truth labels for training inputs. Hence, we use both the answers and rationales output by LLMs as ground truth. In practice, this approach can be applied when it is challenging or costly to obtain ground truth labels. Suppose for the $i$-th input $x_i$, the predicted label from LLM is $\hat y_i$ \footnote{Note that we assign label 1 for 'Yes' answer and label 0 for 'No' answer.}
and the associated rationale is $r_i$. Suppose the student model is $(y_i^{s}, r_i^{s}) = f^s_{\theta}(x_i)$ parameterized by $\theta$. Then the loss function is defined as 
\begin{equation}
    L^s(\theta) = \sum_{i=1}^{D}(CE(y_i^{s}, \hat y_i) + CE(r_i^{s}, r_i)),
    \vspace{-0.02in}
\end{equation}
where $D$ is the total number of training data points and $CE$ represents the cross-entropy averaged at token-level.

\vspace{-0.02in}
\subsubsection{Distillation with Labels.} Assume that we have access to ground truth labels for training inputs. Hence, we can use the ground truth labels for fine-tuning. Specifically, if the LLMs can not predict the right answer, we only use the binary ground truth labels. Otherwise, we use both the labels and rationales. Formally, the loss function is defined as 
\begin{equation}
\label{eq:distill}
    L^s(\theta) = \sum_{i=1}^{D}(CE(y_i^{s}, y_i) + \lambda*\mathbf{1}_{y_i=\hat y_i} CE(r_i^{s}, r_i)),
    \vspace{-0.02in}
\end{equation}
where $\lambda$ is an adjustable parameter (see Appendix \ref{appendix:param} for selecting $\lambda$).

\section{Experimental Setup}
\subsection{Datasets}
We evaluate our approach on three public datasets and an Amazon private dataset.

\subsubsection{Toxigen.} This is an GPT3-generated dataset provided by \citeauthor{hartvigsen2022toxigen}  \shortcite{hartvigsen2022toxigen}, which contains both toxic and benign statements about 13 minority groups. We use their annotated dataset with 8,960 training statements and 940 testing statements in our experiments, where 40\% of statements are toxic. Note that we exclude about 5\% ambiguous statements with toxicity level 3 in experiments, where the scores (i.e. toxicity level) range from 1 to 5 and 3 denotes ambiguity, thus are removed from our experiments.
\subsubsection{SBIC.} This dataset contains 44,671 social media posts from Reddit, Twitter, and hate sites. Each post was annotated by the Social Bias Frames proposed in \cite{sap2019social} to specify whether there exists any social bias and stereotype towards a target group that is toxic. In our experiments, we randomly and uniformly sample 4,000 statements as training dataset and 1,000 statements as testing dataset, 50\% of which are toxic.
\subsubsection{DHate.} This dataset is generated by a human-and-model-in-the-loop process proposed in \cite{vidgen2020learning}. In total, it contains about 40,000 labeled statements, of which 54\% contains hate content. We randomly and uniformly sample 4,000 statements as training dataset and 1,000 statements as testing dataset in our experiments.
\subsubsection{Amazon.} This is a private dataset from Amazon, which contains 8,000 benign statements and 2,000 toxic statements annotated by professional human labelers. This dataset is only used for testing purposes in our experiments due to confidentiality policy.

\vspace{-0.03in}
\subsection{Models and Baselines}
\subsubsection{DToT prompting.} We evaluate the effectiveness of DToT prompting on both \textit{black-box} and \textit{white-box} models, which are defined in Section 3.1. Specifically, we select gpt-3.5-turbo (denoted by ChatGPT \cite{chatgpt}) with 175B parameters as our \textit{black-box} model, and we consider FastChat-T5 (denoted by FC-T5 \cite{zheng2023judging}) with 3B parameters as our \textit{white-box} model. 

Moreover, we compare DToT prompting with three existing baselines: (a) RoBerta model fine-tuned on each dataset, since prior work \cite{hartvigsen2022toxigen} has shown that it can achieve the SOTA performance on Toxigen dataset. (b) CoT prompting, which can be viewed as a special case of DToT prompting without iteratively re-prompting. (c) UniLC prompting proposed by \citeauthor{zhang2023interpretable} (\citeyear{zhang2023interpretable}).

Finally, we compare DToT prompting with its two augmented versions: DToT+FS and DToT+FS+R (see Section 3.2). Note that for DToT+FS, we select $K=3$ positive statements and $K=3$ negative statements that are most semantically similar to the input statement from a development set. To measure the similarity, we use sentence transformer \cite{reimers-2019-sentence-bert} to convert each statement into an embedding, and use the cosine similarity between two statements as a measurement of their semantic similarity. Moreover, for DToT+FS+R, we use the rationales associated with the correct answers generated by ChatGPT as augmentations.

\vspace{-0.03in}
\subsubsection{Model Distillation.} For our main experiments, we select FC-T5 to evaluate the effectiveness of fine-tuning a student LM using rationales generated from LLMs, and we consider two baseline approaches: (a) fine-tuning with labels without rationale, 
and (b) fine-tuning with rationales generated by CoT prompting (which we denote by $\mathbf{R}_{CoT}$). Note that for our approach (which we denote by $\mathbf{R}_{DToT}$), we use the rationales generated via DToT+FS+R prompting as the ground truth during fine-tuning. Furthermore, to investigate how the number of parameters in student LM affects the fine-tuning performance, we use Flan-T5 models with different model size \cite{chung2022scaling} (see Appendix A for more details on why we select these models and hyperparameter setup).

\section{Evaluation Results}

\begin{table*}[!h]
\centering
\begin{tabular}{p{0.07\textwidth}p{0.15\textwidth}|p{0.05\textwidth}p{0.05\textwidth}p{0.05\textwidth}|p{0.05\textwidth}p{0.05\textwidth}p{0.05\textwidth}|p{0.05\textwidth}p{0.05\textwidth}p{0.05\textwidth}}
\hline
\multirow{2}{*}{Model}
&\multirow{2}{*}{Method} 
& \multicolumn{3}{c|}{Toxigen}
& \multicolumn{3}{c|}{SBIC} 
& \multicolumn{3}{c}{DHate} 
\\
~ & ~
& Acc & F1 & AUC
& Acc & F1 & AUC 
& Acc & F1 & AUC \\\hline
RoBerta  & FT	
& 82.45 & 75.35 & 90.24
& 82.80 & 83.96 & 91.82 
& 70.60 & 75.38 & 81.27 \\\hline 
\multirow{4}{*}{FC-T5}  & CoT
& 79.69  & 73.49 &	85.54 
& 63.50 &	71.15 &	67.76 
& 63.40 &	71.09 &	69.13 \\ 
~  & DToT        
& 81.24  & 75.36 &	86.67 
& \textbf{68.80} & \textbf{74.26} &	72.47 
& \textbf{66.80} & \textbf{73.05} &	72.66 \\ 

~  & DToT+FS
& 81.90  & 75.88 &	86.74 
& 68.00  & 73.77 &	\textbf{72.74} 
& 65.40  & 72.23 &	\textbf{73.80} \\ 
~  & DToT+FS+R
& \textbf{82.01}  & \textbf{75.99}  & \textbf{86.94} 
& 68.00 &	73.77 &	72.56  
& 66.50 &	72.87 &	73.37 \\ \hline 
\multirow{4}{*}{ChatGPT}  & CoT 
& 82.71  & 81.29 &	N/A 
& 68.50  & 72.82 &	N/A 
& 65.20  & 72.20 &	N/A \\ 
~  & UniLC        
& 83.30  & 82.73 &	N/A 
& N/A    & N/A   &	N/A 
& N/A    & N/A   &	N/A \\ 
~  & DToT        
& 85.03  & 82.76 &	N/A 
& 71.60  & 74.18 &	N/A 
& 68.20  & 73.92 &	N/A \\ 
~  & DToT+FS
& 86.03  & 83.51 &	N/A 
& 71.70  & 74.62 &	N/A 
& \textbf{69.50}  & \textbf{74.33} &	N/A \\ 
~  & DToT+FS+R
& \textbf{87.03}  & \textbf{85.06} &	N/A 
& \textbf{72.00}  & \textbf{74.91} &	N/A 
& 69.20  & 74.30 &	N/A \\ \hline
\end{tabular}
\caption{Evaluation results of DToT on Toxigen, SBIC, DHate datasets. In ``Method" column, ``FT" stands for fine-tuning on training dataset, ``CoT" refers to CoT prompting, ``DToT" corresponds to DToT prompting, ``DToT+FS" denotes the combination of DToT prompting with few-shot demonstrations, and ``DToT+FS+R" presents the combination of DToT prompting with few-shot demonstrations and rationale augmentations. Due to the lack of output logits, the AUC scores of ChatGPT are populated as ``N/A".}
\vspace{-0.1in}
\label{tab_dtot_full}
\end{table*}

\vspace{-0.0in}
\subsection{Evaluation of DToT}
We start by evaluating DToT prompting to answer the following question.
\vspace{-0.02in}
\subsubsection{Q1: Can DToT prompting enhance the detection performance of LLMs?} 
As shown in Table \ref{tab_dtot_full}, compared with CoT prompting, DToT prompting can enhance the zero-shot learning performance of both \textit{black-box} model and \textit{white-box} model significantly on all three public datasets. 
Specifically, for FC-T5, DToT prompting can increase the accuracy by up to 5.30\% and the F1 score by up to 3.11\%. It is worth noting that DToT prompting can also improve the AUC score of FC-T5 on all datasets, indicating its robust performance. For ChatGPT, DToT prompting consistently outperforms CoT prompting, and increases the accuracy by up to 3.10\% and the F1 score by up to 1.72\%. 

Moreover, combining DToT with few-shot in-context learning (i.e. DToT+FS and DToT+FS+R) may further improve models' performance. For instance, on Toxigen dataset, compared with the vanilla DToT prompting, adding demonstrations (i.e. DToT+FS) can improve 
ChatGPT's accuracy by 1.00\%. Furthermore, by incorporating both demonstrations and rationales during prompting (i.e. DToT+FS+R), 
ChatGPT's accuracy can be improved by 2.00\% respectively, outperforming baselines by up to 4.58\% (for RoBerta) and at least 3.73\% (for UniLC).

Therefore, we conclude that DToT prompting and its augmented versions significantly enhance LLMs' performance on toxic content detection.

\begin{table*}[!t]
\centering
\begin{tabular}{p{0.06\textwidth}p{0.07\textwidth}p{0.1\textwidth}|p{0.05\textwidth}p{0.05\textwidth}p{0.05\textwidth}|p{0.05\textwidth}p{0.05\textwidth}p{0.05\textwidth}|p{0.05\textwidth}p{0.05\textwidth}p{0.05\textwidth}}
\hline
\multirow{2}{*}{Model} & \multirow{2}{*}{Label} &\multirow{2}{*}{Rationale} 
& \multicolumn{3}{c|}{Toxigen} 
& \multicolumn{3}{c|}{SBIC} 
& \multicolumn{3}{c}{DHate}\\
~ & ~ & ~
& Acc & F1 & AUC
& Acc & F1 & AUC 
& Acc & F1 & AUC \\\hline
RoBerta  & Human & N/A
& 82.45 & 75.35 & 90.24
& 82.80 & 83.96 & 91.82 
& 70.60 & 75.38 & 81.27 \\\hline 
ChatGPT  & N/A & N/A
& 87.03 & 85.06 & N/A
& 72.00 & 74.91 & N/A 
& 69.50 & 74.33 & N/A \\\hline 
\multirow{6}{*}{FC-T5}  & \multirow{3}{*}{LLM} & N/A
& 81.90  & 80.19 &	91.78  
& 64.90  & 71.99 & 74.67 
& 63.50  & 71.15  & 70.50 \\ 
~ & ~ & $\mathbf{R}_{CoT}$ 
& 81.79  & 80.10 &	92.88 
& 67.30  & 73.35 &	74.25 
& 64.30  & 71.60 &	72.90 \\ 

~ & ~ & $\mathbf{R}_{DToT}$
& \textbf{84.00} & \textbf{82.08} & \textbf{93.60}  
& \textbf{69.00} & \textbf{74.42} &	\textbf{81.09} 
& \textbf{68.00} & \textbf{73.77} &	\textbf{77.89} \\ \cline{2-12} 
~ & \multirow{3}{*}{Human} & N/A
& 84.99  & 83.00 &	92.43  
& 84.00  & 84.91 &	92.89 
& 85.00  & 85.71 & 93.64 \\ 
~ & ~ & $\mathbf{R}_{CoT}$ 
& 87.31  & 85.24 &	94.15  
& 84.20  & 85.07 &	93.18 
& 86.20  & 86.71 &  93.71 \\ 
~ & ~ & $\mathbf{R}_{DToT}$
& \textbf{87.53} & \textbf{85.46} & \textbf{94.37}  
& \textbf{85.10} & \textbf{85.82} &	\textbf{93.85} 
& \textbf{86.40} & \textbf{86.87}  & \textbf{94.49} \\ \hline 
\end{tabular}
\caption{Distillation evaluation results on Toxigen, SBIC, and DHate datasets. In ``Label" column, ``Human" indicates that the labels come from the training dataset, ``LLM" indicates that the labels are predicted by LLM. In ``Rationale" column,  ``N/A" means no rationales are used in fine-tuning, ``$\mathbf{R}_{CoT}$" means rationales extracted via CoT are used in fine-tuning, and ``$\mathbf{R}_{CoT}$" means rationales extracted via DToT are used in fine-tuning.}
\vspace{-0.1in}
\label{tab_distill_v0}
\end{table*}

\subsection{Evaluation of Rationale Distillation}
Next, we evaluate whether fine-tuning with rationales extracted via DToT improves the performance of student LMs. We first answer the following question:

\subsubsection{Q2: Can fine-tuning with rationales extracted via DToT derive a student LM with higher accuracy?} 
Table \ref{tab_distill_v0} reports the evaluation results of fine-tuning with or without rationales. In this table,
``Label = Human" denotes the utilization of the ground truth labels in training datasets for fine-tuning, ``Label = LLM" means the usage of labels predicted by the teacher LLM for fine-tuning, ``Rationale = N/A" indicates that fine-tuning is conducted solely with labels,  ``Rationale = $\mathbf{R}_{CoT}$" implies that fine-tuning takes place with both labels and CoT-extracted rationale from teacher LLMs, and  ``Rationale = $\mathbf{R}_{DToT}$" represents that we fine-tune the model with both labels and DToT-extracted rationales from the teacher LLM (i.e. ChatGPT). 

As reported in Table \ref{tab_distill_v0}, FC-T5 fine-tuned with DToT-extracted rationales and ground truth labels outperforms all baselines across various public datasets. Notably, this fine-tuning approach yields significant improvements in accuracy, F1 score, and AUC score. Specifically, compared with fine-tuning with labels only, it can increase the model accuracy by up to 2.54\%, the F1 score by up to 2.46\%, and the AUC score by up to 1.94\% respectively.  Compared with the prior SOTA LM fine-tuned on these datasets (i.e. RoBerta), it can significantly increase the model accuracy by up to 15.80\%, the F1 score by up to 11.49\%, and the AUC score by up to 13.22\% respectively. In addition, FC-T5 fine-tuned with labels and rationales outperforms teacher LLM by up to 16.90\% w.r.t. accuracy, with 60$\times$ smaller model size.

Moreover, compared with fine-tuning with $\mathbf{R}_{CoT}$ and ground truth labels, we observe that fine-tuning with $\mathbf{R}_{DToT}$ and ground truth labels can consistently result in student LMs with better detection performance on all public datasets. This indicates that rationales generated via DToT prompting ($\mathbf{R}_{DToT}$) have higher quality than those generated via CoT prompting ($\mathbf{R}_{CoT}$). 

Lastly, under the scenario where we use labels predicted by teacher LLM as ground truth, we observe that fine-tuning with $\mathbf{R}_{DToT}$ and labels can still outperform both fine-tuning with $\mathbf{R}_{CoT}$ and labels, and fine-tuning with only labels. Specifically, on Toxigen dataset, FC-T5 fine-tuned with $\mathbf{R}_{DToT}$ and predicted labels from teacher LLM can even outperform RoBerta fine-tuned with human labels from training dataset. However, without using $\mathbf{R}_{DToT}$, the fine-tuned FC-T5 cannot outperform Roberta.

In summary, we conclude that fine-tuning with both labels and rationales can effectively improve the student LMs' performance. Moreover, using rationales with better quality (i.e. DToT-extracted rationales versus CoT-extracted rationales) can further enhance the performance of fine-tuned LMs.

\begin{table}[!t]
\centering
\begin{tabular}{p{0.06\textwidth}p{0.05\textwidth}p{0.07\textwidth}|p{0.05\textwidth}p{0.05\textwidth}p{0.05\textwidth}}

\hline
\multirow{1}{*}{Model} & \multirow{1}{*}{Labels} &\multirow{1}{*}{Rationale} 
& \multicolumn{1}{c}{SBIC} 
& \multicolumn{1}{c}{DHate} 
& \multicolumn{1}{c}{Amazon}\\\hline
RoBerta  & Human & N/A
& 65.46  
& 61.51 
& X \\\hline
ChatGPT  & N/A & N/A
& 72.00  
& 69.50 
& N/A \\\hline
\multirow{6}{*}{FC-T5}  & \multirow{3}{*}{LLM} & N/A
& 75.36   
& 74.96   
& X+2.47 \\ 
~ & ~ & $\mathbf{R}_{CoT}$
& 71.48  
& 70.99 
& X+4.52 \\ 
~ & ~ & $\mathbf{R}_{DToT}$
& \textbf{75.90} 
& \textbf{77.31} 
& \textbf{X+4.54} \\ \cline{2-6} 
~ & \multirow{3}{*}{Human} & N/A
& 75.24 
& 77.75  
& X+0.05 \\ 
~ & ~ & $\mathbf{R}_{CoT}$
& 77.15  
&\textbf{78.97} 
& \textbf{X+4.16} \\ 
~ & ~ & $\mathbf{R}_{DToT}$
& \textbf{77.29} 
& 77.54 
& X+3.61 \\ \hline 
\end{tabular}
\vspace{-0.03in}
\caption{Transferability evaluation results. Note that we fine-tune these models on Toxigen dataset while testing them on other datasets, and we report AUC score. For Amazon dataset, due to confidentiality policy, we only report the increased AUC score compared with RoBerta (whose AUC score is denoted by X). }
\vspace{-0.12in}
\label{tab_transfer_v0}
\end{table}

\subsubsection{Q3: Can fine-tuning with rationales improve the transferability of student LMs across different toxic datasets?} 
To evaluate whether fine-tuning with both labels and rationales can improve the transferability of student LMs, we test the models fine-tuned on the other two public datsaets (SBIC and DHate) and one private dataset (Amazon). Table \ref{tab_transfer_v0} reports our transferability evaluation results, where we use AUC score as our metric. 
First, we observe that 
compared with fine-tuning with labels only (Rationale = N/A), fine-tuning with both labels and rationales (Rationale = $\mathbf{R}_{CoT}$/$\mathbf{R}_{DToT}$) can improve the AUC score of student LMs by up to 2.35\% on testing datasets. Second, FC-T5 fined-tuned with both labels and rationales have significantly better AUC scores on all datasets compared with RoBerta. Lastly, while we only fine-tune FC-T5 on Toxigen datset, it still outperforms teacher LLM (i.e. ChatGPT) on both SBIC and DHate datasets. Hence, we conclude that fine-tuning with rationale effectively improves the cross-dataset transferability of student LMs.

\subsubsection{Q4: How does student LMs' size affect the detection accuracy?}
\begin{table}[!h]
\centering
\begin{tabular}{p{0.07\textwidth}p{0.04\textwidth}p{0.08\textwidth}|p{0.04\textwidth}p{0.04\textwidth}p{0.04\textwidth}}
\hline
Model & Size & Rationale & Acc & F1 & AUC \\\hline
\multirow{2}{*}{FC-T5} & \multirow{2}{*}{3B} & N/A
& 84.99  & 83.00 &	92.43 \\
~  & ~ & $\mathbf{R}_{DToT}$
& \textbf{87.53}  & \textbf{85.46}  & \textbf{94.37} \\\hline
\multirow{2}{*}{F-T5-XL} & \multirow{2}{*}{3B} & N/A
& 85.54  & 83.52 &	92.51 \\ 
~  & ~ & $\mathbf{R}_{DToT}$
& 87.53  & 85.42 &	93.17 \\\hline 

\multirow{2}{*}{F-T5-L} & \multirow{2}{*}{770M} & N/A
& 83.44  & 81.57 &	92.42\\ 
~  & ~ & $\mathbf{R}_{DToT}$
& 84.22  & 82.24 &	93.18 \\\hline 

\multirow{2}{*}{F-T5-B} & \multirow{2}{*}{220M} & N/A
& 81.02  & 79.43 &	91.26 \\ 
~  & ~ & $\mathbf{R}_{DToT}$ 
& 81.35  & 79.66 &	91.16 \\ 
\hline 
\end{tabular}
\vspace{-0.03in}
\caption{Impact of student LMs' size on rationale distillation. Note that we use Toxigen dataset for a case study, and F-T5-XL/F-T5-L/F-T5-B is short for Flan-T5-XL/Flan-T5-Large/Flan-T5-Base.}
\vspace{-0.1in}
\label{tab_size}
\end{table}
So far, we use FC-T5 with 3B parameters as our student LM during rationale distillation experiments. To investigate how the model size affects the student LMs' performance, we fine-tune Flan-T5 models of different size with rationales on Toxigen dataset. As reported in Table \ref{tab_size}, fine-tuning with both labels and rationales can consistently enhance the student LMs' performance on Toxigen dataset with varying model size. Moreover, as the model size becomes smaller, the model performance will decrease, and the performance gain provided by using rationales also becomes smaller, which indicates that larger student LMs can learn to generate rationales better.

\subsubsection{Q5: Can student LMs fine-tuned with rationales actually generate high-quality rationales?}
We conduct a case study on the quality of rationales generated by FC-T5 using different approaches via manual check. We observe that after fine-tuning with both labels and rationales ($\mathbf{R}_{DToT}$), FC-T5 always provides responses with rich rationales.
In contrast, fine-tuning with labels only makes FC-T5 overfit the binary labels and hence output `Yes/No' answers only without rationales.
In addition, without fine-tuning, FC-T5 cannot consistently generate meaningful rationales (see Table \ref{tab:r_quality} in Appendix C for detailed examples). 

\section{Conclusions and Limitations}
In this work, we propose an end-to-end approach to bootstrapping and distilling LLMs for toxic content detection, where a novel prompting method named DToT is designed to enhance LLMs' detection performance and extract better rationales, and smaller LMs are fine-tuned with both labels and DToT-extracted rationales. Our evaluation results on four datasets consistently demonstrate the effectiveness of both the proposed DToT prompting and the proposed fine-tuning method for toxic content detection.
\subsubsection{Limitations.} First, the context selector of DToT conducts greedy search at each iteration, which may not be globally optimal. Second, we use a pre-defined context-tree in our experiments for DToT, which can not dynamically change with LLMs' response.

\appendix
\section{Appendix}
\subsection{Experimental Parameters and Models}
\label{appendix:param}
\subsubsection{Parameters for DToT Prompting.} We set $s_l$ and $s_h$ in Eq. (\ref{eq:score}) as 0 and 90 respectively for experiments with ChatGPT. Since we observe that ChatGPT will answer 'Yes' as long as the toxicity rating is above zero, which leads to high recall while relatively low precision (i.e. high false positive), we select a small value for $s_l$ and large value for $s_h$ to reduce the false positive rate of ChatGPT. We set $s_{\delta}$ in Eq. (\ref{eq:conf}) as 0.9 for confidence checker in DToT prompting, since we empirically notice that setting the threshold of confidence score as 0.9 improves the confidence of generated answers and thus enhances the overall detection accuracy.

Moreover, we use a two-level context tree in our experiments (as shown in Figure \ref{fig:tree} in Section 3.1) and the maximal terminal step $T$ is set as 2. Note that we did not experiment with a deeper context-tree since DToT prompting with a two-level context tree can already significantly improve LLMs' detection performance with reasonable cost.

\subsubsection{Parameters for Rationale Distillation.} We set $\lambda$ in Eq. (\ref{eq:distill}) as 1, in order to make the predicted answers and rationales of student LMs equally important in our experiments. During rationale distillation experiments (see Section 5.2), for each dataset, we fine-tune the student LM on the training set for 5 epochs, where stochastic gradient descent with mini-batch size of 8 and learning rate of 1e-4 are used.

\subsubsection{Models for DToT Prompting.} We select ChatGPT as our \textit{black-box} LLM since it is the most popular model with SOTA performance. Specifically, we use gpt-3.5-turbo instead of gpt-4, since gpt-3.5-turbo has much better trade-off between model performance and experimental cost. For \textit{white-box} LLMs, we have tried different open-source LLMs including FastChat-T5 \cite{zheng2023judging}, Flan-T5-XXL \cite{chung2022scaling}, and Falcon-40B \cite{falcon40b}. We finally select FastChat-T5 since it has the best detection accuracy using prompts from Table \ref{tab:tab_dtot}.

\subsubsection{Models for Distillation.} We use FastChat-T5 \cite{zheng2023judging} as the student LM since it has the best performance after fine-tuning, compared with other models from T5 family \cite{chung2022scaling} (see Table \ref{tab_size} in Section 5.2). We do not consider models larger than 3B during distillation experiments, due to their high fine-tuning cost while low run-time efficiency.

\subsection{Rationale Examples}
\label{appendix:r_eg}
Table \ref{tab:r_quality} demonstrates the responses generated by FC-T5 using different approaches, where the input sentence is \textit{``why do you have to come to America, you should stay in your own country and fight for your freedom, not come to America"}. We observe that after fine-tuning with both labels and rationales (FT+$\mathbf{R}_{DToT}$), FC-T5 can provide a response with rich rationale (see ``Rationale: The sentence ..."). By contrast, fine-tuning with only labels will make FC-T5 output a `Yes/No' answer without any rationale. Without fine-tuning, FC-T5 also fails to output a reasonable rationale that explains why this sentence contains toxic content. 

\begin{table}[!h]
\centering
\begin{tabular}{p{0.06\textwidth}p{0.1\textwidth}p{0.23\textwidth}}
\hline
Model & Method &  Response \\\hline
\multirow{3}{*}{FC-T5}  & CoT 
& Yes, this sentence contains toxic content.
\\\cline{2-3}
~  & FT 
& Yes.\\\cline{2-3}
~  & FT+$\mathbf{R}_{DToT}$  
& Yes, this sentence contains toxic content. Rationale: The sentence contains hate speech and biased content towards immigrants. It implies that immigrants are not welcome in America and should stay in their own country to fight for their freedom.
\\ \hline
\end{tabular}
\caption{Responses generated by FC-T5 using different approaches. Note we require to detect toxic content in sentence \textit{``why do you have to come to America, you should stay in your own country and fight for your freedom, not come to America"} in the prompt.}
\vspace{-0.1in}
\label{tab:r_quality}
\end{table}

\subsection{Prompt Templates}
\label{appendix:prompt}
Table \ref{tab:tab_dtot} below provides the prompt templates used in our experiments, where the question $q$ in Section 3.1 is given as ``Does sentence \texttt{s} contain toxic content?". Note that for black-box LLM (ChatGPT), we explicitly ask the model to rate the toxicity level during prompting, in order to calculate the confidence score (see Section 3.1 Confidence Checker for details).
\begin{table}[h!]
\centering
\begin{tabular}{p{0.1\textwidth}p{0.3\textwidth}}
\hline
Model & Prompt template \\\hline
ChatGPT  & \makecell[l]{
Context: \texttt{c}. \\
Sentence: \texttt{s}. \\
Does this sentence contain \texttt{type(c)} \\
content?\\
First, answer in ``Yes" or ``No". \\
Second, rate the \texttt{type(c)} level out of 100.\\
Third, explain your rationale briefly. \\
The output format is given below: \\
Answer: ...\\
\texttt{type(c)} level: .../100.\\
Rationale: ...}
\\ \hline
FC-T5  & \makecell[l]{
\texttt{c}. \\
\#\#\# Human: ``\texttt{s}". Does this sentence \\
contain \texttt{type(c)} content? \\
Answer yes or no, and explain your answer.\\
\#\#\# Assistant:}
\\\hline
\end{tabular}
\caption{Prompt templates for different LLMs, given question $q$: ``Does sentence \texttt{s} contain toxic content?" (see Section 3.1). Note that \texttt{c} is the context, \texttt{s} is the statement, and \texttt{type(c)}  is the category of context \texttt{c} (e.g. toxic, hate and violent).} 
\label{tab:tab_dtot}
\end{table}
\setcounter{secnumdepth}{0}
\section{Acknowledgments}
The authors would like to thank anonymous reviewers and members from Amazon for providing insightful feedback. This work is supported in part by the National Science Foundation under grant numbers 1956435 and 1901488.
\bibliography{aaai24}

\end{document}